\documentclass[letterpaper, 10 pt, conference]{ieeeconf}  

\IEEEoverridecommandlockouts                              

\title{\LARGE \bf
Advancing On-Device Neural Network Training with TinyPropv2: Dynamic, Sparse, and Efficient Backpropagation
}

\author{Marcus Rüb$^{1}$, Axel Sikora$^{1}$ and Daniel Mueller-Gritschneder$^{2}$
\thanks{$^{1}$Software Solutions, Hahn-Schickard, Villingen-Schwenningen, Germany
        {\tt\small firstname.name@hahn-schickard.de}}%
\thanks{$^{2}$Electronic Design Automation, Technical University of Munich, Munich, Germany 
        {\tt\small Daniel.Mueller@tum.de}}%
}
\usepackage{amsmath,amssymb,amsfonts}
\usepackage{algorithmic}
\usepackage{float} 
\usepackage{color}
\usepackage[dvipsnames,table,xcdraw]{xcolor}
\usepackage{pgfplots}
\usepackage{algorithm}
\usepackage{algorithmic}

\usepackage{colortbl}
\usepackage{hhline}

\usepackage{listings}
\usepackage{booktabs} 
\usepackage{graphicx} 
\newcommand{\highlight}[1]{\cellcolor{gray!25}{#1}}

\definecolor{backcolour}{rgb}{0.85,0.85,0.85}
\usepackage[sorting=none]{biblatex} 
\def\BibTeX{{\rm B\kern-.05em{\sc i\kern-.025em b}\kern-.08em
    T\kern-.1667em\lower.7ex\hbox{E}\kern-.125emX}}

\begin{document}

\maketitle
\thispagestyle{empty}
\pagestyle{empty}

\begingroup
\let\clearpage\relax

\begin{abstract}
This study introduces TinyPropv2, an innovative algorithm optimized for on-device learning in deep neural networks, specifically designed for low-power microcontroller units. TinyPropv2 refines sparse backpropagation by dynamically adjusting the level of sparity, including the ability to selectively skip training steps. This feature significantly lowers computational effort without substantially compromising accuracy. Our comprehensive evaluation across diverse datasets—CIFAR 10, CIFAR100, Flower, Food, Speech Command, MNIST, HAR, and DCASE2020—reveals that TinyPropv2 achieves near-parity with full training methods, with an average accuracy drop of only around 1\% in most cases. For instance, against full training, TinyPropv2's accuracy drop is minimal, for example, only 0.82\% on CIFAR 10 and 1.07\% on CIFAR100. In terms of computational effort, TinyPropv2 shows a marked reduction, requiring as little as 10\% of the computational effort needed for full training in some scenarios, and consistently outperforms other sparse training methodologies. These findings underscore TinyPropv2's capacity to efficiently manage computational resources while maintaining high accuracy, positioning it as an advantageous solution for advanced embedded device applications in the IoT ecosystem.

\end{abstract}

\section{Introduction}

Deep learning has revolutionized the landscape of machine learning and artificial intelligence, enabling significant advancements across numerous applications such as image recognition, natural language processing, and autonomous systems. Central to this revolution is the ability to train deep neural networks (DNNs) effectively. However, as network architectures become deeper and datasets grow larger, the computational burden of training these models using traditional backpropagation algorithms has surged, often outstripping the capabilities of low-power, embedded devices. \cite{RUB2022272}

Embedded systems, particularly microcontroller units (MCUs), are ubiquitous in the Internet of Things (IoT) applications, where on-device learning offers a plethora of benefits, including privacy preservation, reduced latency, and decreased reliance on continuous cloud connectivity. However, the limited computational and memory resources of such devices pose a significant challenge for deploying sophisticated DNNs.

In response to this challenge, sparse backpropagation algorithms have emerged as an attractive solution, optimizing the training process by selectively updating a subset of the model's weights. Yet, the static nature of the backpropagation ratio in existing approaches often results in a precarious balance between computational efficiency and model accuracy. 

Building on previous results TinyProp \cite{Rub.18.08.2023}, this paper introduces TinyPropv2, an enhanced algorithm that dynamically adjusts the backpropagation ratio during the training process.
TinyPropv2 extends this dynamic adaptability further by incorporating a decision-making process that can skip entire training steps for certain datapoints when they are deemed unnecessary, thereby reducing computational effort without significantly impacting accuracy. Through rigorous experimentation and analysis, we demonstrate that TinyPropv2 not only conserves computational resources but also provides a safeguard against overfitting, thus representing a significant step forward in the quest for efficient and effective on-device learning.

This introduction provides an overview of the challenges in on-device learning and positions TinyPropv2 as a novel contribution by reduce the computational demands of backpropagation.  
\\
The remainder of the paper is structured as follows: Section II reviews related work and contextualizes TinyPropv2 within the landscape of sparse backpropagation methods. Section III details the methodology underlying TinyPropv2, elucidating its innovative approach to adaptive backpropagation. Section IV presents the experimental setup and datasets utilized, ensuring reproducibility and a comprehensive understanding of the evaluation context. Section V discusses the results, highlighting the accuracy and computational efficiency of TinyPropv2 across various datasets. Finally, Section VI concludes the paper and outlines directions for future work, underscoring the potential impact of TinyPropv2 on the broader domain of machine learning and embedded systems.

\begin{figure*}[!hbt] 
	\centering
	\includegraphics[width=0.9\textwidth]{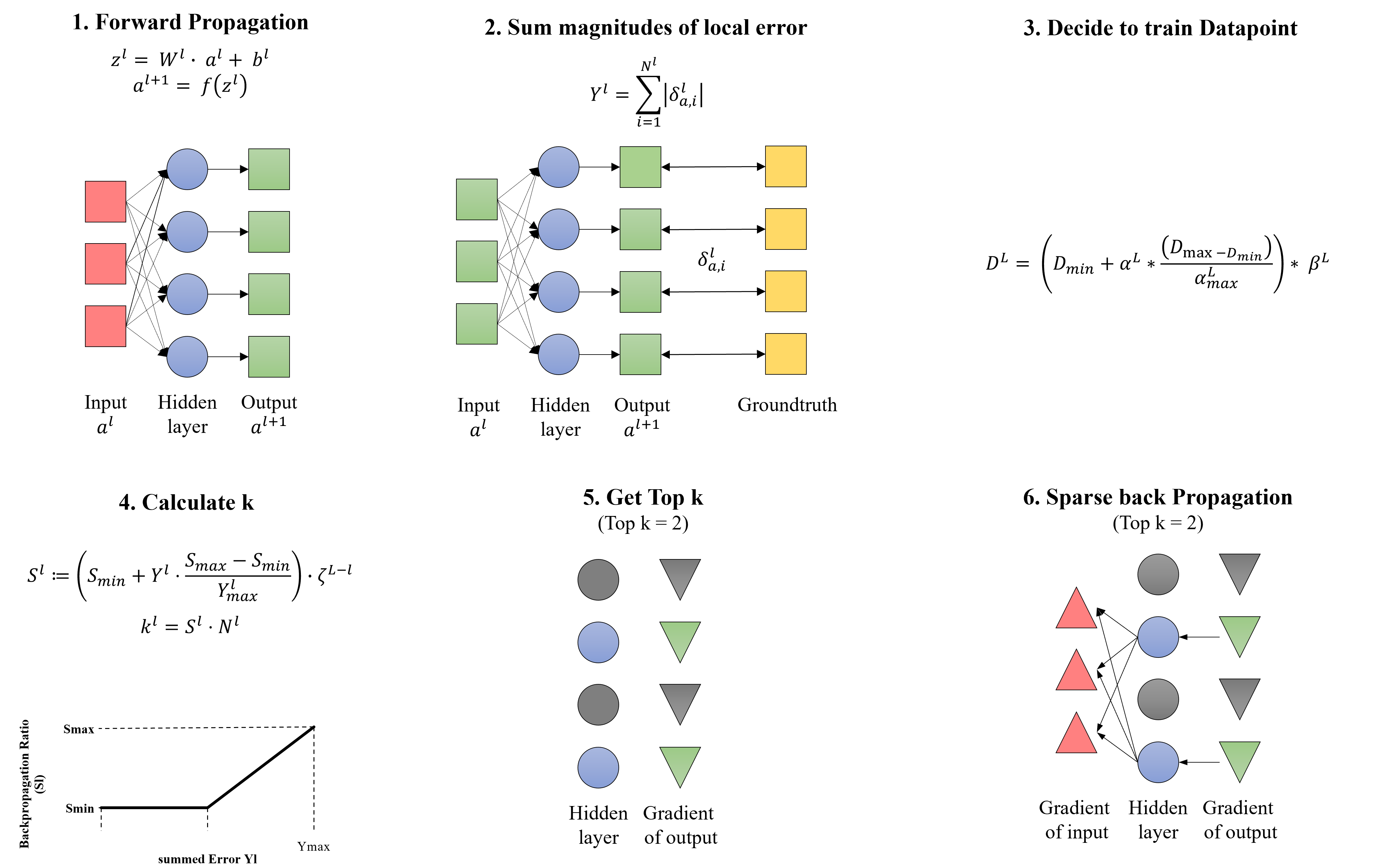} 
	\caption{Operational Workflow of TinyPropv2: The process begins with (1) performing the forward pass to compute the output. This is followed by (2) calculating the loss function and accumulating the local errors. Subsequently, (3) a decision is made on whether to train the datapoint based on the computed error. Next, (4) the optimal number of gradients to update, denoted as 'local k,' is determined from the aggregated error. (5) The algorithm then identifies the top 'k' gradients that will be updated. Finally, (6) these selected gradients undergo the sparse backpropagation process, completing the training step.}
	\label{tinyprop}
\end{figure*}

\section{related work}

The evolution of on-device learning, a cornerstone in IoT applications, pivots around the inefficiencies of traditional offline training and deployment models, which often fail to adapt to real-time data distribution shifts~\cite{Spadaro.10220231062023}. Continual learning offers a solution by focusing on acquiring knowledge step-by-step, much like how humans learn, while also addressing the issue of catastrophic forgetting, where a model loses previously learned information~\cite{Goodfellow.2015,Kirkpatrick.2017,Parisi.2019}.

A critical challenge in this field is the limitation of computational and memory resources on embedded devices, especially during backpropagation. Approaches to mitigate these challenges bifurcate into enhancing architecture efficiency and implementing sparse updates. Sparse updates, particularly, have gained attention for reducing the memory footprint during backpropagation by selectively updating network layers based on various criteria~\cite{Zhu.2023,Goli.2020,MahdiNikdan.2023,Liu.02.10.2023}.

The landscape of on-device learning research can be broadly categorized into:

\begin{itemize}
    \item \textbf{Adapting Various Machine Learning Algorithms for Embedded Devices:} Research efforts like those of Lee et al.~\cite{Lee.1212201612142016} and SEFR~\cite{Keshavarz.08.06.2020} have explored retraining edge-level ML algorithms and implementing low-power classifiers, respectively. However, these studies primarily do not focus on neural networks.
    \item \textbf{Training DNNs on Embedded Devices:} Techniques such as TinyOL~\cite{Ren.15.03.2021} and TinyTL~\cite{tinytl} have been developed for fine-tuning DNNs, but their scope is limited and does not encompass full neural network training.
    \item \textbf{Sparse Backpropagation Variations:} This area includes innovations like meProp~\cite{Sun.20.06.2017} and various top-k sparsity methods~\cite{Wei.18.09.2017,Zhu.01.06.2018,Chmiel.15.06.2020}. These methods are focused on optimizing backpropagation by sparsifying the gradient vector, yet they suffer from limitations like fixed sparsity levels.
\end{itemize}

In this context, TinyPropv2 introduces a dynamic approach, differing from static methods like meprop ~\cite{Sun.17.11.2017}. Unlike other techniques that assume fixed top-k sparsity or a log-normal distribution in gradients~\cite{Chmiel.15.06.2020}, TinyPropv2 dynamically adjusts the number of gradients to update based on error propagation rates and local errors.

Recent advancements, such as TinyTrain~\cite{Kwon.19.07.2023} and the approach in~\cite{Yang.uuuuuuuu}, highlight other dimensions of on-device learning. TinyTrain reduces training time through task-adaptive sparse updates, and~\cite{Yang.uuuuuuuu} introduces strategies for dynamic neuron update selection under memory constraints. These differ from TinyPropv2’s focus on optimizing DNN training efficiency through adaptive sparse backpropagation.

In conclusion, TinyPropv2 uniquely addresses the limitations in existing on-device learning methods. It optimizes for computational resources on constrained devices by dynamically adjusting backpropagation based on error rates, setting it apart from other methods in its adaptability and efficiency.

\section{Methodology}

Deep learning, especially in the context of deep neural networks (DNNs), has shown remarkable success in various applications. However, the training process, which involves forward and backward propagation, can be computationally intensive. This section introduces the standard forward and backward propagation methods and explains the concept of sparse backpropagation, the basis for the original TinyProp algorithm. We then detail the advancements made in our enhanced version, TinyPropv2.

\subsection{Forward and Backward Propagation}

Deep Neural Networks (DNNs) primarily consist of a series of layers where each layer transforms its input data to produce an output. This transformation is achieved through a combination of linear and non-linear operations. The process of computing the output of the network given an input is termed as forward propagation. Conversely, the process of updating the network's weights based on the error of its predictions is termed as backward propagation. \cite{Rumelhart.1986}

\subsubsection{Forward Propagation}

For a given layer \( l \), the forward propagation can be mathematically represented as:

\begin{equation}
a^{l+1} = f\left(z^{l}\right) = f\left(W^{l} a^{l} + b^{l}\right)
\end{equation}

where:
\begin{itemize}
    \item \( W^{l} \) is the weight matrix for layer \( l \).
    \item \( b^{l} \) is the bias vector for layer \( l \).
    \item \( z^{l} \) represents the weighted sum of inputs for layer \( l \).
    \item \( a^{l} \) is the activation (output) of layer \( l \).
    \item \( f \) is a non-linear activation function.
\end{itemize}

\subsubsection{Backward Propagation}

Once the network produces an output, the error or loss is computed. This loss is then used to update the weights of the network to improve its predictions. The process of computing the gradient of the loss with respect to the network's weights and biases is termed as backward propagation.

The loss \( \mathcal{L} \) is given by:

\begin{equation}
\mathcal{L}(a^{L},y)
\end{equation}

where \( y \) is the ground truth and \( a^{L} \) is the output of the final layer \( L \).

The gradient of the loss with respect to the pre-activation \( z^{L} \) of the last layer is:

\begin{equation}
\delta_{z}^{L} = \left( \frac{\partial \mathcal{L}}{\partial a^{L}} \frac{\partial a^{L}}{\partial z^{L}} \right)^T = {f}'\left(z^{L}\right) \odot \nabla_{a^L} \mathcal{L}
\end{equation}

where \( f' \) is the derivative of the activation function and \( \odot \) represents the Hadamard (element-wise) product.

\subsection{Sparse Backpropagation}

Sparse backpropagation is an optimization technique that aims to reduce the computational complexity of the standard backpropagation algorithm. Instead of updating all the weights in the network, sparse backpropagation updates only a subset of them, specifically those with the largest gradients. This approach is based on the observation that only a few weights, which have the most significant gradients, contribute the most to the learning process. \cite{Sun.20.06.2017}

\subsubsection{Gradient Sparsity}

Given a gradient vector \( \delta_{a}^{l} \) for layer \( l \), the top-k elements based on their magnitude can be represented as:

\begin{equation}
\text{top}(\delta_{a}^{l}, k)
\end{equation}

For instance, for a gradient vector \( v = [1, 2, 3, -4]^T \), the top-2 elements would be represented as \( \text{top}(v, 2) = [0, 0, 3, -4]^T \).

\subsubsection{Approximate Gradient}

The approximate gradient is then computed by retaining only the top-k elements and setting the rest to zero:

\begin{equation}
\hat{\delta}^{l}_{a} = 
\begin{cases} 
\delta^{l}_{a} & \text{if } a \in \{t_{1}, t_{2}, \ldots, t_{k}\} \\
0 & \text{otherwise}
\end{cases}
\end{equation}

This approximation ensures that only the most significant gradients contribute to the weight updates, leading to a reduction in computational effort.

\subsubsection{Sparse Gradient Propagation}

Using the approximate gradient, the backpropagation is modified as:

\begin{align}
\hat{\delta}^{l}_{a} &= \text{top}(\delta_{a}^{l}, k) \\
\hat{\delta_{z}^{l}} &= \hat{\delta_{a}^{l}} \odot f'\left(z^{l}\right) \\
\hat{\delta^{l-1}_{a}} &= W^{l-1} \hat{\delta}^{l}_{z}
\end{align}

Where \( f' \) is the derivative of the activation function.

By employing this sparse approach, the backpropagation algorithm becomes more efficient, especially for deep networks with a large number of parameters.

\subsection{The TinyProp Algorithm}

The TinyProp algorithm enhances the efficiency of training deep neural networks (DNNs) by implementing a dynamic sparse backpropagation approach. This method is particularly effective for on-device learning on tiny, embedded devices such as low-power microcontroller units (MCUs). \cite{Rub.18.08.2023}

\subsubsection{TinyProp Adaptivity Approach}

The core innovation of TinyProp lies in its adaptivity, where the algorithm dynamically calculates the proportion of gradients to be updated for each layer during training. This adaptivity is based on the local error in each layer, enabling the algorithm to focus computational resources more effectively.

\paragraph{Local Error Vector and Total Error}

The adaptivity mechanism in TinyProp uses the local error vector \(\delta ^{l}_{a,i}\) of each layer to gauge the layer's contribution to the overall error. The total error characteristic of the layer \(Y^{l}\) is computed as the sum of the absolute values of these error components:

\begin{equation}
Y^{l} = \sum\limits_{i=1}^{N^{l}} \left\lvert \delta ^{l}_{a,i} \right\rvert
\end{equation}

\paragraph{Adaptive Error Propagation Rate}

TinyProp introduces an adaptive error propagation rate \(S^l\), which reflects the training progress and is calculated as a function of the total error:

\begin{equation}
S^{l} = S_{\text{min}} + Y^{l} \frac{S_{\text{max}} - S_{\text{min}}}{Y^{l}_{\text{max}}}
\end{equation}

where \(S_{\text{max}}\) and \(S_{\text{min}}\) are user-defined bounds for the error propagation rate.

\paragraph{TinyProp Damping Factor}

To address the computational intensity in larger DNN layers, TinyProp incorporates a damping factor \(\zeta(l)\) that reduces the error propagation rate in a layer-dependent manner:

\begin{equation}
S^{l} = \left(S_{\text{min}} + Y^{l} \frac{S_{\text{max}} - S_{\text{min}}}{Y^{l}_{\text{max}}}\right) \zeta^{-l+L}
\end{equation}

This factor allows for reduced computational effort in layers that typically require less training, such as the initial layers of the network.

\paragraph{Computing the Adaptive Top-k}

With the calculated error propagation rate \(S^l\), TinyProp determines the number of gradients to update in each layer adaptively:

\begin{equation}
k^{l} = S^{l} \cdot N^{l}
\end{equation}

\paragraph{TinyProp Backpropagation Algorithm}

The backpropagation step in TinyProp is then conducted using the adaptive top-k approach:

\begin{equation}
\begin{aligned}
\hat{\delta} ^{l}_{a} &= \text{top}(\delta_{a}^{l}, k^{l})\\
\hat{\delta_{z}^{l}} &= \hat{\delta_{a}^{l}} \odot {f}'\left(z^{l}\right)\\
\hat{\delta ^{l-1}_{a}} &= \left(W^{l-1}\right)^{T}\cdot \hat{\delta} ^{l}_{z}
\end{aligned}
\end{equation}

Through this methodology, TinyProp efficiently manages computational resources while maintaining the efficacy of the learning process, making it highly suitable for embedded applications.

\subsection{The Enhanced TinyPropv2 Algorithm}

Building upon the original TinyProp algorithm, TinyPropv2 introduces an additional layer of decision-making to potentially skip entire training steps when beneficial, see Fig. \ref{tinyprop}. The primary motivation for incorporating this additional decision layer stems from the need to enhance computational efficiency without sacrificing the model's accuracy. Traditional training methods often expend significant computational resources on processing every data point, regardless of its actual impact on the model's learning. TinyPropv2, by contrast, intelligently identifies and focuses on data points that substantially contribute to the learning process. This targeted approach ensures that computational efforts are allocated more judiciously, leading to a more efficient training cycle. This approach further reduces computational effort while maintaining the balance between efficiency and accuracy.

\subsubsection{Decision Mechanism for Training Data Points}

The decision to train a specific data point in TinyPropv2 is based on a novel decision metric, \( D \), which assesses the necessity of performing backpropagation for that data point:

\begin{equation}
D^{L} = \left( D_{\text{min}} + \alpha^{L} \frac{D_{\text{max}} - D_{\text{min}}}{\alpha^{L}_{\text{max}}} \right) \times \beta^{L}
\end{equation}

Where:
\begin{itemize}
    \item \( D^{L} \) is the decision metric for the last layer \( L \).
    \item \( D_{\text{min}} \) and \( D_{\text{max}} \) are the minimum and maximum thresholds for the decision metric.
    \item \( \alpha^{L} \) is a factor based on the current state of training at layer \( L \).
    \item \( \beta^{L} \) is a scaling factor to adjust the sensitivity of the decision metric.
\end{itemize}

If \( D^{L} \) exceeds a certain threshold, such as 0.5, backpropagation is performed; otherwise, it is skipped. This selective approach prioritizes data points that are more impactful for learning, enhancing efficiency.

\paragraph{Threshold Determination:}
The threshold against which \( D^{L} \) is compared is not arbitrarily set but is carefully chosen based on empirical observations and domain-specific requirements. For instance, a threshold value of 0.5 is commonly used as a starting point. However, this threshold can be adjusted according to the characteristics of the dataset and the specific learning goals. 

\begin{itemize}
    \item \textbf{Empirical Tuning:} The threshold is often empirically tuned during the preliminary stages of model training. This involves experimenting with different threshold values and observing their impact on the model's performance and training efficiency.
    \item \textbf{Dataset Sensitivity:} The optimal threshold may vary depending on the dataset's complexity and the nature of the data points. For example, datasets with more noise or variability might require a different threshold approach compared to more uniform datasets.
    \item \textbf{Performance Metrics:} The decision on the threshold also considers the balance between training efficiency and accuracy. A higher threshold might speed up training but at the cost of accuracy, and vice versa.
\end{itemize}

\subsubsection{Benefits of TinyPropv2 Over TinyProp}

TinyPropv2 extends the capabilities of the original TinyProp algorithm by:
\begin{itemize}
    \item \textbf{Reducing Computational Load:} By intelligently deciding whether to perform backpropagation for each data point.
    \item \textbf{Enhanced Adaptability:} Offers more refined control over the training process, suitable for various types of datasets and learning scenarios.
    \item \textbf{Resource Optimization:} Particularly beneficial for MCUs and embedded devices where computational resources are limited.
\end{itemize}

\subsection{Pseudocode for the TinyPropv2 Algorithm}

The following pseudocode outlines the steps involved in the TinyProp and TinyPropv2 algorithms, highlighting the dynamic sparse backpropagation approach and the decision mechanism unique to TinyPropv2.

\begin{algorithm}
\caption{TinyPropv2 Algorithm}
\begin{algorithmic}[1]
\REQUIRE Training data, Network architecture
\ENSURE Trained network weights
\FOR{each training epoch}
    \FOR{each data point}
        \STATE \textbf{/* TinyPropv2 Decision Mechanism */}
        \STATE Compute decision metric $D^{L}$ (TinyPropv2)
        \IF{$D^{L}$ exceeds threshold (TinyPropv2)}
            \STATE \textbf{/* End of TinyPropv2-specific step */}
            \FOR{each layer $l$ in the network}
                \STATE Compute forward propagation
                \STATE Calculate local error vector $\delta^{l}_{a,i}$
                \STATE Compute total layer error $Y^{l}$
                \STATE Calculate adaptive error propagation rate $S^{l}$
                \STATE Compute damping factor adjustment $\zeta(l)$
                \STATE Determine adaptive top-k value $k^{l}$
                \STATE Compute sparse gradient $\hat{\delta}^{l}_{a}$
            \ENDFOR
            \FOR{each layer $l$ in backward order}
                \STATE Apply sparse backpropagation updates
            \ENDFOR
        \ELSE
            \STATE Skip backpropagation for this data point (TinyPropv2)
        \ENDIF
    \ENDFOR
\ENDFOR
\end{algorithmic}
\end{algorithm}

\begin{table*}[htbp]
\centering
\caption{Accuracy of different methods across various datasets.}
\label{tab:accuracy}
\begin{tabular}{lcccccccc}
\hline
\textbf{Method} & \textbf{CIFAR 10} & \textbf{CIFAR100} & \textbf{Flower} & \textbf{Food} & \textbf{Speech Command} & \textbf{MNIST} & \textbf{HAR} & \textbf{DCASE} \\
\hline
\textbf{Model} & MobileNetV2 & MobileNetV2 & MobileNetV2 & MobileNetV2 & 5 L DNN & 5 L DNN & 5 L DNN & 5 L DNN \\
Full Training & 96.12 & 80.9 & 94.01 & 80.4 & 96.4 & 96.6 & 95.3 & 98.88 \\
Sparse Update & 95.13 & 78.6 & \highlight{93.77} & 77.81 & 93.98 & 96.41 & 94.3 & 97.17 \\
Velocity & 95.25 & 79.46 & 93.03 & 79.16 & 94.51 & 96.44 & 94.5 & 97.96 \\
TinyTrain & 94.91 & 79.51 & 93.33 & 79.23 & 94.6 & \highlight{96.53} & 95.1 & 97.97 \\
TinyProp & 93.8 & 78.6 & 91.6 & 78.3 & 93.0 & 96.32 & 94.1 & 97.55 \\
TinyPropv2 & \highlight{95.3} & \highlight{79.83} & 93.7 & \highlight{79.1} & \highlight{95.3} & \highlight{96.53} & \highlight{95.2} & \highlight{98.23} \\
\hline
\end{tabular}
\end{table*}

\section{Experiment Setup}

\subsection{Datasets Utilized}

The experimental validation of the TinyPropv2 algorithm was conducted on a heterogeneous set of open-source datasets, chosen for their prevalence in benchmarking within the machine learning community as well as their representation of diverse application domains.

\paragraph{CIFAR-10 and CIFAR-100} 
The CIFAR-10 and CIFAR-100 datasets are staples in image classification challenges. CIFAR-10 comprises 60,000 32x32 color images evenly distributed across 10 classes, while CIFAR-100 shares the same total number of images but is spread thinly over 100 classes, increasing the granularity and complexity of the classification task. The uniformity of image size and pre-established splits for training and testing make these datasets ideal for evaluating model generalizability and robustness. \cite{Krizhevsky09learningmultiple}

\paragraph{Oxford Flowers 102} 
The Oxford Flowers 102 dataset, a collection of 8,189 images, is categorized into 102 flower classes that vary significantly in scale, pose, and light conditions. The dataset poses a fine-grained visual classification challenge due to the subtle differences between classes and significant intra-class variations. \cite{Nilsback.1216200812192008}

\paragraph{Food-101} 
Comprising 101 food categories totaling 101,000 images, the Food-101 dataset presents a real-world scenario of noisy data. The training set is intentionally uncleaned to simulate the practical challenges encountered in image recognition tasks, with the added difficulty of dealing with imbalanced datasets due to the variability in the number of images per category. \cite{bossard14}

\paragraph{Speech Commands} 
This dataset encompasses 65,000 one-second long audio clips of 30 different spoken words by various speakers. The dataset provides a rigorous test bed for speech recognition models, challenging them to distinguish between nuanced audio signals. \cite{speechcommandsv2}

\paragraph{MNIST} 
MNIST, a classic dataset in the machine learning field, consists of 70,000 28x28 grayscale images of handwritten digits. It serves as a benchmark for evaluating the performance of image processing systems. \cite{lecun2010mnist}

\paragraph{Human Activity Recognition (HAR)} 
The HAR dataset captures daily activities of 30 subjects via waist-mounted smartphones. It contains time-series data derived from accelerometer and gyroscope sensors, presenting a challenge in activity recognition from multi-dimensional time-series sensor data. \cite{GarciaGonzalez.2020}

\paragraph{DCASE2020 Challenge Task 1} 
Selected from the DCASE2020 Challenge, this dataset contains recordings from six different machine types. For this study, we focused on the slide rail machine type, using normal operational sounds to create a binary classification task indicative of normal and anomalous machine behaviors. \cite{Dcase.01.02.2022}

Each dataset was selected not only for its individual complexity but also for the collective breadth they provide, encompassing a wide array of challenges including image recognition, audio processing, and sensor data analysis. This diversity ensures a rigorous validation of the TinyPropv2 method across various data types and real-world scenarios.

\subsection{Models and Architectures}
Our experiments leveraged the MobileNetV2 architecture \cite{Sandler.2018}, renowned for its efficiency on mobile devices, and a bespoke 5-layer neural network tailored to the dataset modality—employing either 1D or 2D convolutions for time-series and image data, respectively. These models were initially pretrained on the ImageNet dataset to establish a foundational knowledge before being fine-tuned for our specific datasets.

\subsection{Computational Environment}

A critical aspect of evaluating the efficacy of any machine learning algorithm, particularly those designed for on-device deployment, is the computational environment in which the experiments are performed. For our experiments, we selected a controlled computational setting that would reflect the constraints and capabilities of high-performance embedded systems.

\paragraph{Software Framework} 
We utilized Python as the primary programming language for its widespread adoption in the scientific computing community and its comprehensive suite of machine learning libraries. The experiments were implemented using the PyTorch 2.0.0 framework, benefiting from its dynamic computation graph and efficient memory management for deep neural network training.

\paragraph{Training Protocol} 
Consistency in the training protocol was maintained across all experiments to ensure comparability. We adopted Stochastic Gradient Descent (SGD) \cite{H.Robbins.1951} without momentum or weight decay, coupled with a cosine annealing scheduler to modulate the learning rate across 200 epochs. The learning rate was initialized at 0.125 and annealed to zero, with a warm-up phase of 5 epochs. This training policy was selected to mirror typical fine-tuning practices in resource-limited settings, where complex adaptive optimization algorithms may not be feasible.

\paragraph{Evaluation Metric} 
Accuracy was determined by evaluating the top-1 performance metric on the respective test sets (D\textsubscript{test}) of each dataset. This metric was chosen for its straightforward interpretability and its common use as a benchmark in classification tasks.

The computational environment was carefully architected to strike a balance between replicating the limitations of embedded devices and ensuring the reproducibility of results. It provided a rigorous testbed for our algorithms and a reliable indicator of their potential performance in real-world applications.

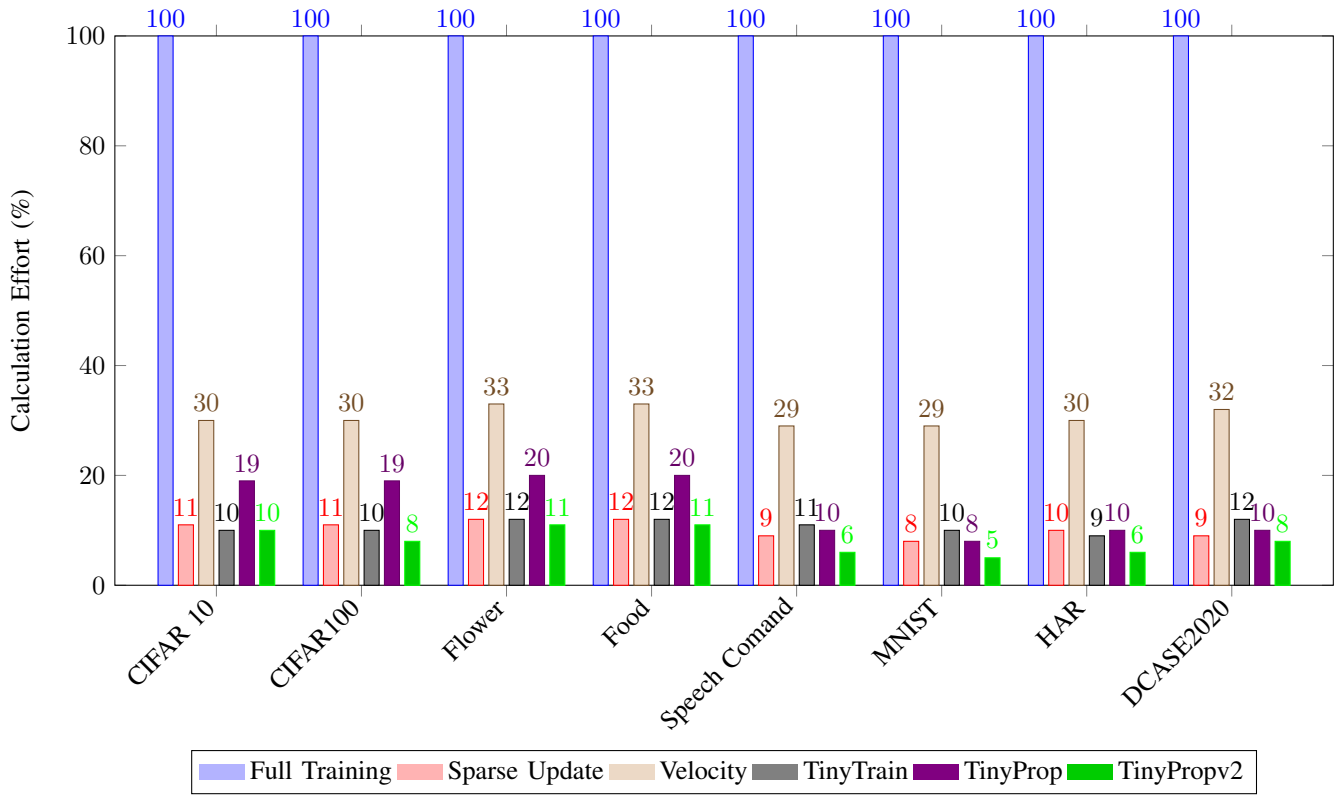
\begin{figure*}[htbp]
\centering
\begin{tikzpicture}
    \begin{axis}[
        ybar,
        bar width=.2cm,
        width=\textwidth,
        height=.5\textwidth,
        legend style={at={(0.5,-0.3)},
          anchor=north,legend columns=-1},
        symbolic x coords={CIFAR 10,CIFAR100,Flower,Food,Speech Comand,MNIST,HAR,DCASE2020},
        xtick=data,
        nodes near coords,
        nodes near coords align={vertical},
        ymin=0,ymax=100,
        ylabel={Calculation Effort (\%)},
        x tick label style={rotate=45,anchor=east},
        legend image code/.code={
            \draw[#1, draw=none] (0cm,-0.1cm) rectangle (0.6cm,0.1cm);
        },
        ]
        \addplot coordinates {(CIFAR 10,100) (CIFAR100,100) (Flower,100) (Food,100) (Speech Comand,100) (MNIST,100) (HAR,100) (DCASE2020,100)};
        \addplot coordinates {(CIFAR 10,11) (CIFAR100,11) (Flower,12) (Food,12) (Speech Comand,9) (MNIST,8) (HAR,10) (DCASE2020,9)};
        \addplot coordinates {(CIFAR 10,30) (CIFAR100,30) (Flower,33) (Food,33) (Speech Comand,29) (MNIST,29) (HAR,30) (DCASE2020,32)};
\addplot coordinates {(CIFAR 10,10) (CIFAR100,10) (Flower,12) (Food,12) (Speech Comand,11) (MNIST,10) (HAR,9) (DCASE2020,12)};
\addplot coordinates {(CIFAR 10,19) (CIFAR100,19) (Flower,20) (Food,20) (Speech Comand,10) (MNIST,8) (HAR,10) (DCASE2020,10)};
\addplot coordinates {(CIFAR 10,10) (CIFAR100,8) (Flower,11) (Food,11) (Speech Comand,6) (MNIST,5) (HAR,6) (DCASE2020,8)};
    \legend{Full Training,Sparse Update,Velocity,TinyTrain,TinyProp,TinyPropv2}
    
\end{axis}
\end{tikzpicture}
\caption{Comparative analysis of computational effort required for different training methods across various datasets.}
\label{bar_chart}
\end{figure*}

\section{Results}

The evaluation of the TinyPropv2 method involved a dual-focus analysis: first, we assessed the accuracy of the method across various datasets; second, we examined the computational effort required during training. This two-pronged approach enabled us to investigate not only the effectiveness of the model in terms of learning capabilities but also its efficiency, which is crucial for deployment on resource-constrained devices.

\subsection{Accuracy}

Our accuracy assessment revealed that TinyPropv2 competently navigated the trade-off between model complexity and learning accuracy shown in Tab. \ref{tab:accuracy}. In datasets with higher-dimensional data and more complex structures, such as CIFAR-100 and DCASE2020, TinyPropv2 demonstrated a remarkable capacity to maintain high accuracy levels, rivaling the full training baseline without necessitating extensive computational resources. 

In simpler datasets like MNIST, the accuracy advantage of TinyPropv2 over other methods became less pronounced, suggesting that its benefits are more significant in scenarios where the learning task inherently involves more complexity and where discerning the salient features from the data is more challenging.

\subsection{Computational Effort}

One of the notable features of TinyPropv2 is its ability to skip training on certain datapoints as the model becomes more competent. This approach effectively combats overfitting by reducing the training intensity as the model's accuracy improves. Consequently, as the model training progresses and the algorithm identifies less error-prone data, the computational effort required diminishes significantly.

The computational effort analysis, as depicted in the accompanying bar chart, shows that TinyPropv2 rapidly decreases its computational load relative to other methods. Initially, the effort aligns closely with that of Sparse Update and TinyTrain approaches. However, as training progresses and the algorithm becomes more selective in the datapoints it deems necessary to train on, we observe a steeper decline in computational effort for TinyPropv2.

This behavior underscores the method's adaptability and responsiveness to the learning progress, offering an efficient training process that dynamically adjusts to the model's evolving state of knowledge. It affirms TinyPropv2's potential as a scalable solution for on-device learning, where computational resources are often at a premium and must be judiciously allocated.

\subsection{Comparative Analysis}

The comparative analysis of computational effort required for different training methods is illustrated in Figure 2. As shown, TinyPropv2 starts on par with other methods but as training continues, the effort required for TinyPropv2 decreases more steeply. This is due to its unique ability to skip redundant datapoint training, thereby reducing unnecessary computations. As a result, TinyPropv2 not only conserves computational resources but also mitigates the risk of overtraining, striking a desirable balance between learning efficiency and model performance.

\subsection{Implications for On-device Learning}

The implications of these results are significant for on-device machine learning applications. TinyPropv2's intelligent computation management makes it particularly suited for environments where power and processing capabilities are limited. By minimizing computational overhead without compromising on learning outcomes, TinyPropv2 ensures that devices such as smartphones, IoT sensors, and other embedded systems can perform complex learning tasks autonomously.

The results from this study provide a compelling case for the adoption of TinyPropv2 in on-device learning scenarios. Its dynamic adjustment to the training process not only optimizes the computational load but also enhances the overall longevity and functionality of devices operating in resource-constrained environments.

\section{Conclusions and Future Work}

The comprehensive experimental analysis conducted in this study leads us to several important conclusions about the TinyPropv2 algorithm's capabilities and potential applications. TinyPropv2 demonstrates a notable advancement in on-device learning, offering a method that judiciously uses computational resources while still achieving high levels of accuracy across a range of datasets and learning tasks.

\subsection{Conclusions}

Our results confirm that TinyPropv2 can effectively reduce the computational effort required during the training process by selectively skipping datapoints that do not contribute significantly to model improvement. This strategic reduction in training intensity does not only conserve energy and computational resources but also presents a lessened risk of overfitting, a common pitfall in machine learning endeavors. 

TinyPropv2's performance was particularly impressive in complex and high-dimensional datasets, where it successfully approached the upper accuracy limits set by full training baselines. This finding underscores the algorithm's suitability for complex learning tasks that are characteristic of real-world applications, ranging from image and speech recognition to sensor data analysis.

\subsection{Future Work}

The promising results obtained with TinyPropv2 open several avenues for future research. One immediate direction is the exploration of TinyPropv2's performance on an even wider array of datasets, including those with unstructured data or in unsupervised learning settings. Additionally, further optimization of the algorithm's hyperparameters could yield even more efficient training processes and higher accuracies.

Another important area of future work involves the deployment of TinyPropv2 on actual embedded systems. Real-world testing will provide invaluable insights into the algorithm's performance in the field and its interaction with hardware limitations. 

Furthermore, extending TinyPropv2 to support federated learning environments could significantly enhance its utility. In such settings, the algorithm's efficiency in computation and communication would be paramount, enabling robust learning across distributed devices with minimal data exchange.

Lastly, the integration of reinforcement learning principles could provide mechanisms to further refine the decision-making process behind the selective updating of the model. This would allow TinyPropv2 to dynamically adapt to changing environments and tasks, making it even more versatile and powerful for on-device learning applications.

\subsection{Implications}

The implications of this research are twofold: not only does it contribute to the theoretical understanding of efficient on-device learning, but it also provides a practical framework that can be readily applied in various industries. From consumer electronics to industrial IoT, the applications of TinyPropv2 are vast and impactful, making it a significant contribution to the field of machine learning.

In conclusion, TinyPropv2 stands as a highly effective tool for on-device machine learning, balancing the dual demands of computational efficiency and learning accuracy. Its continued development and adaptation will undoubtedly contribute to the advancement of edge computing and the realization of truly intelligent devices.

\endgroup

\addtolength{\textheight}{-12cm}   




\AtNextBibliography{\footnotesize}
\printbibliography

@article{RUB2022272,
title = {A Practical View on Training Neural Networks in the Edge},
journal = {IFAC-PapersOnLine},
volume = {55},
number = {4},
pages = {272-279},
year = {2022},
note = {17th IFAC Conference on Programmable Devices and Embedded Systems PDES 2022 — Sarajevo, Bosnia and Herzegovina, 17-19 May 2022},
issn = {2405-8963},
doi = {https://doi.org/10.1016/j.ifacol.2022.06.045},
url = {https://www.sciencedirect.com/science/article/pii/S2405896322003603},
author = {Marcus Rüb and Prof. Dr. Axel Sikora}
}

@inproceedings{Nilsback.1216200812192008,
 author = {Nilsback, Maria-Elena and Zisserman, Andrew},
 title = {Automated Flower Classification over a Large Number of Classes},
 pages = {722--729},
 publisher = {IEEE},
 booktitle = {2008 Sixth Indian Conference on Computer Vision, Graphics {\&} Image Processing},
 year = {12/16/2008 - 12/19/2008},
 doi = {10.1109/ICVGIP.2008.47}
}

@article{lecun2010mnist,
  title={MNIST handwritten digit database},
  author={LeCun, Yann and Cortes, Corinna and Burges, CJ},
  journal={ATT Labs [Online]. Available: http://yann.lecun.com/exdb/mnist},
  volume={2},
  year={2010}
}

@article{GarciaGonzalez.2020,
 author = {Garcia-Gonzalez, Daniel and Rivero, Daniel and Fernandez-Blanco, Enrique and Luaces, Miguel R.},
 year = {2020},
 title = {A Public Domain Dataset for Real-Life Human Activity Recognition Using Smartphone Sensors},
 volume = {20},
 number = {8},
 journal = {Sensors (Basel, Switzerland)},
 doi = {10.3390/s20082200}
}

@article{speechcommandsv2,
   author = { {Warden}, P.},
    title = "{Speech Commands: A Dataset for Limited-Vocabulary Speech Recognition}",
  journal = {ArXiv e-prints},
  archivePrefix = "arXiv",
  eprint = {1804.03209},
  primaryClass = "cs.CL",
    year = 2018,
    month = apr,
    url = {https://arxiv.org/abs/1804.03209},
}

@inproceedings{bossard14,
  title = {Food-101 -- Mining Discriminative Components with Random Forests},
  author = {Bossard, Lukas and Guillaumin, Matthieu and Van Gool, Luc},
  booktitle = {European Conference on Computer Vision},
  year = {2014}
}

@inproceedings{Krizhevsky09learningmultiple,
            author={Alex Krizhevsky},
            title={Learning multiple layers of features from tiny images},
            institution={},
            year={2009}
}

@inproceedings{Sandler.2018,
 author = {Sandler, Mark and Howard, Andrew and Zhu, Menglong and Zhmoginov, Andrey and Chen, Liang-Chieh},
 title = {MobileNetV2: Inverted Residuals and Linear Bottlenecks},
 pages = {4510--4520},
 publisher = {IEEE},
 isbn = {978-1-5386-6420-9},
 booktitle = {2018 IEEE/CVF Conference on Computer Vision and Pattern Recognition (CVPR 2018)},
 year = {2018},
 address = {Piscataway, NJ},
 doi = {10.1109/CVPR.2018.00474}
}

@article{H.Robbins.1951,
 author = {{H. Robbins}},
 year = {1951},
 title = {A Stochastic Approximation Method},
 journal = {Annals of Mathematical Statistics}
}

@article{Rumelhart.1986,
 author = {Rumelhart, David E. and Hinton, Geoffrey E. and Williams, Ronald J.},
 year = {1986},
 title = {Learning representations by back-propagating errors},
 url = {https://www.nature.com/articles/323533a0},
 pages = {533--536},
 volume = {323},
 number = {6088},
 issn = {1476-4687},
 journal = {Nature},
 doi = {10.1038/323533a0}
}

@inproceedings{Spadaro.10220231062023,
 author = {Spadaro, Gabriele and Renzulli, Riccardo and Bragagnolo, Andrea and Giraldo, Jhony H. and Fiandrotti, Attilio and Grangetto, Marco and Tartaglione, Enzo},
 title = {Shannon Strikes Again! Entropy-based Pruning in Deep Neural Networks for Transfer Learning under Extreme Memory and Computation Budgets},
 pages = {1510--1514},
 publisher = {IEEE},
 isbn = {979-8-3503-0744-3},
 booktitle = {2023 IEEE/CVF International Conference on Computer Vision Workshops (ICCVW)},
 year = {10/2/2023 - 10/6/2023},
 doi = {10.1109/ICCVW60793.2023.00165}
}

@misc{Rub.18.08.2023,
 author = {R{\"u}b, Marcus and Maier, Daniel and Mueller-Gritschneder, Daniel and Sikora, Axel},
 date = {18.08.2023},
 title = {TinyProp -- Adaptive Sparse Backpropagation for Efficient TinyML  On-device Learning},
 url = {http://arxiv.org/pdf/2308.09201.pdf}
}

@article{Parisi.2019,
 author = {Parisi, German I. and Kemker, Ronald and Part, Jose L. and Kanan, Christopher and Wermter, Stefan},
 year = {2019},
 title = {Continual lifelong learning with neural networks: A review},
 url = {https://www.sciencedirect.com/science/article/pii/S0893608019300231},
 pages = {54--71},
 volume = {113},
 journal = {Neural networks : the official journal of the International Neural Network Society},
 doi = {10.1016/j.neunet.2019.01.012}
}

@article{MahdiNikdan.2023,
 author = {{Mahdi Nikdan} and {Tommaso Pegolotti} and {Eugenia Iofinova} and {Eldar Kurtic} and {Dan Alistarh}},
 year = {2023},
 title = {SparseProp: Efficient Sparse Backpropagation for Faster Training of Neural Networks at the Edge},
 url = {https://proceedings.mlr.press/v202/nikdan23a.html},
 pages = {26215--26227},
 issn = {2640-3498},
 journal = {International Conference on Machine Learning}
}

@misc{Liu.02.10.2023,
 author = {Liu, Liyuan and Gao, Jianfeng and Chen, Weizhu},
 date = {02.10.2023},
 title = {Sparse Backpropagation for MoE Training},
 url = {http://arxiv.org/pdf/2310.00811.pdf}
}

@misc{Kwon.19.07.2023,
 author = {Kwon, Young D. and Li, Rui and Venieris, Stylianos I. and Chauhan, Jagmohan and Lane, Nicholas D. and Mascolo, Cecilia},
 date = {19.07.2023},
 title = {TinyTrain: Deep Neural Network Training at the Extreme Edge},
 url = {http://arxiv.org/pdf/2307.09988.pdf}
}

@inproceedings{Yang.uuuuuuuu,
 author = {Yang, Yuedong and Li, Guihong and Marculescu, Radu},
 title = {Efficient On-Device Training via Gradient Filtering},
 pages = {3811--3820},
 publisher = {IEEE},
 isbn = {979-8-3503-0129-8},
 booktitle = {2023 IEEE/CVF Conference on Computer Vision and Pattern Recognition (CVPR)},
 year = {uuuu-uuuu},
 doi = {10.1109/CVPR52729.2023.00371}
}

@article{Kirkpatrick.2017,
 author = {Kirkpatrick, James and Pascanu, Razvan and Rabinowitz, Neil and Veness, Joel and Desjardins, Guillaume and Rusu, Andrei A. and Milan, Kieran and Quan, John and Ramalho, Tiago and Grabska-Barwinska, Agnieszka and Hassabis, Demis and Clopath, Claudia and Kumaran, Dharshan and Hadsell, Raia},
 year = {2017},
 title = {Overcoming catastrophic forgetting in neural networks},
 pages = {3521--3526},
 volume = {114},
 number = {13},
 journal = {Proceedings of the National Academy of Sciences of the United States of America},
 doi = {10.1073/pnas.1611835114}
}

@inproceedings{Goli.2020,
 author = {Goli, Negar and Aamodt, Tor M.},
 title = {ReSprop: Reuse Sparsified Backpropagation},
 url = {https://openaccess.thecvf.com/content_CVPR_2020/html/Goli_ReSprop_Reuse_Sparsified_Backpropagation_CVPR_2020_paper.html},
 pages = {1548--1558},
 year = {2020}
}

@book{Goodfellow.2015,
 author = {Goodfellow, Ian J. and Mirza, Mehdi and {Da Xiao} and Courville, Aaron and Bengio, Yoshua},
 year = {2015},
 title = {An Empirical Investigation of Catastrophic Forgetting in Gradient-Based  Neural Networks}
}

@inproceedings{Zhu.2023,
 author = {Zhu, Ligeng and Hu, Lanxiang and Lin, Ji and Chen, Wei-Ming and Wang, Wei-Chen and Gan, Chuang and Han, Song},
 title = {PockEngine: Sparse and Efficient Fine-tuning in a Pocket},
 pages = {1381--1394},
 publisher = {{Association for Computing Machinery}},
 isbn = {9798400703294},
 series = {ACM Digital Library},
 booktitle = {Proceedings of the 56th Annual IEEE/ACM International Symposium on Microarchitecture},
 year = {2023},
 address = {[Erscheinungsort nicht ermittelbar]},
 doi = {10.1145/3613424.3614307}
}

@misc{Wei.18.09.2017,
 author = {Wei, Bingzhen and Sun, Xu and Ren, Xuancheng and Xu, Jingjing},
 date = {18.09.2017},
 title = {Minimal Effort Back Propagation for Convolutional Neural Networks}
}

@misc{Sun.17.11.2017,
 author = {Sun, Xu and Ren, Xuancheng and Ma, Shuming and Wei, Bingzhen and Li, Wei and Xu, Jingjing and Wang, Houfeng and Zhang, Yi},
 date = {2020},
 title = {Training Simplification and Model Simplification for Deep Learning: A  Minimal Effort Back Propagation Method},
 number = {2},
 doi = {10.1109/TKDE.2018.2883613}
}

@misc{Sun.20.06.2017,
 author = {Sun, Xu and Ren, Xuancheng and Ma, Shuming and Wang, Houfeng},
 date = {20.06.2017},
 title = {meProp: Sparsified Back Propagation for Accelerated Deep Learning with  Reduced Overfitting}
}

@misc{Ren.15.03.2021,
 author = {Ren, Haoyu and Anicic, Darko and Runkler, Thomas},
 date = {15.03.2021},
 title = {TinyOL: TinyML with Online-Learning on Microcontrollers}
}

@misc{Keshavarz.08.06.2020,
 author = {Keshavarz, Hamidreza and Abadeh, Mohammad Saniee and Rawassizadeh, Reza},
 date = {08.06.2020},
 title = {SEFR: A Fast Linear-Time Classifier for Ultra-Low Power Devices}
}

@inproceedings{Lee.1212201612142016,
 author = {Lee, Jongmin and Stanley, Michael and Spanias, Andreas and Tepedelenlioglu, Cihan},
 title = {Integrating machine learning in embedded sensor systems for Internet-of-Things applications},
 pages = {290--294},
 publisher = {IEEE},
 booktitle = {2016 IEEE International Symposium on Signal Processing and Information Technology (ISSPIT)},
 year = {12/12/2016 - 12/14/2016},
 doi = {10.1109/ISSPIT.2016.7886051}
}

@misc{Zhu.01.06.2018,
 author = {Zhu, Maohua and Clemons, Jason and Pool, Jeff and Rhu, Minsoo and Keckler, Stephen W. and Xie, Yuan},
 date = {01.06.2018},
 title = {Structurally Sparsified Backward Propagation for Faster Long Short-Term  Memory Training}
}

@misc{Dcase.01.02.2022,
 author = {Dcase},
 year = {01.02.2022},
 title = {DCASE2020 Challenge - DCASE},
 url = {http://dcase.community/challenge2020/index},
 urldate = {01.02.2022}
}

@misc{mnist,
 year = {09.04.2019},
 title = {LeCun et al. (1999): The MNIST Dataset Of Handwritten Digits (Images) --- PyMVPA 2.6.5.dev1 documentation},
 url = {http://www.pymvpa.org/datadb/mnist.html},
 urldate = {07.02.2022}
}

@misc{tinytl,
 author = {Cai, Han and Gan, Chuang and Zhu, Ligeng and Han, Song},
 date = {22.07.2020},
 title = {TinyTL: Reduce Activations, Not Trainable Parameters for Efficient  On-Device Learning}
}

@misc{Chmiel.15.06.2020,
 author = {Chmiel, Brian and Ben-Uri, Liad and Shkolnik, Moran and Hoffer, Elad and Banner, Ron and Soudry, Daniel},
 date = {15.06.2020},
 title = {Neural gradients are near-lognormal: improved quantized and sparse  training},
 url = {https://arxiv.org/pdf/2006.08173}
}





\end{document}